\pdfoutput=1

\documentclass[11pt]{article}

\usepackage[]{acl}

\usepackage{times}
\usepackage{latexsym}

\usepackage[T1]{fontenc}

\usepackage[utf8]{inputenc}

\usepackage{microtype}

\usepackage{graphicx}

\usepackage{color}

\usepackage{bm}

\usepackage{amssymb,amsmath}
\usepackage{amsfonts} 
\usepackage{bbm}
\usepackage{colortbl}
\usepackage{subcaption}

\title{Average Token Delay: A Latency Metric for Simultaneous Translation}

\author{Yasumasa Kano \quad Katsuhito Sudoh \quad Satoshi Nakamura \\
  Nara Institute of Science and Technology, Japan\\
  \texttt{kano.yasumasa.kw4@is.naist.jp}}

\begin{document}
\maketitle

\begin{abstract}
Simultaneous translation is a task in which translation begins before the speaker has finished speaking.
In its evaluation, we have to consider the latency of the translation in addition to the quality.
The latency is preferably as small as possible for users to comprehend what the speaker says with a small delay.
Existing latency metrics focus on when the translation starts but do not consider adequately when the translation ends.
This means such metrics do not penalize the latency caused by a long translation output, which actually delays users' comprehension.
In this work, we propose a novel latency evaluation metric called \emph{Average Token Delay} (ATD) that focuses on the end timings of partial translations in simultaneous translation.
We discuss the advantage of ATD using simulated examples and also investigate the differences between ATD and Average Lagging with simultaneous translation experiments.
\end{abstract}

\section{Introduction}\label{sec:introduction}
Machine translation (MT) has evolved rapidly using recent neural network techniques and is now widely used both for written and spoken languages.
MT for speech is a very attractive application for translating various conversations, lecture talks, etc.
For smooth real-time speech communication across languages, speech MT should run in real-time and incrementally without waiting for the end of an input utterance as in consecutive interpretation.
While the translation quality can improve by waiting for later inputs as the context, it should result in longer latency.
This quality-latency trade-off is the most important issue in simultaneous MT.

Most recent simultaneous MT studies use BLEU \citep{papineni-etal-2002-bleu} for evaluating the quality and Average Lagging \citep[AL;][]{ma-etal-2019-stacl} for the latency.
AL is based on the number of input words that have become available when starting a translation and measures its average over all the timings of generating partial translations.
It is very suitable for wait-$k$ \citep{ma-etal-2019-stacl} that waits for $k$ input tokens before starting the translation and then repeats to read and write one token alternately.

\begin{figure}[t]
\centering
\centerline{\includegraphics[width=7.5cm]{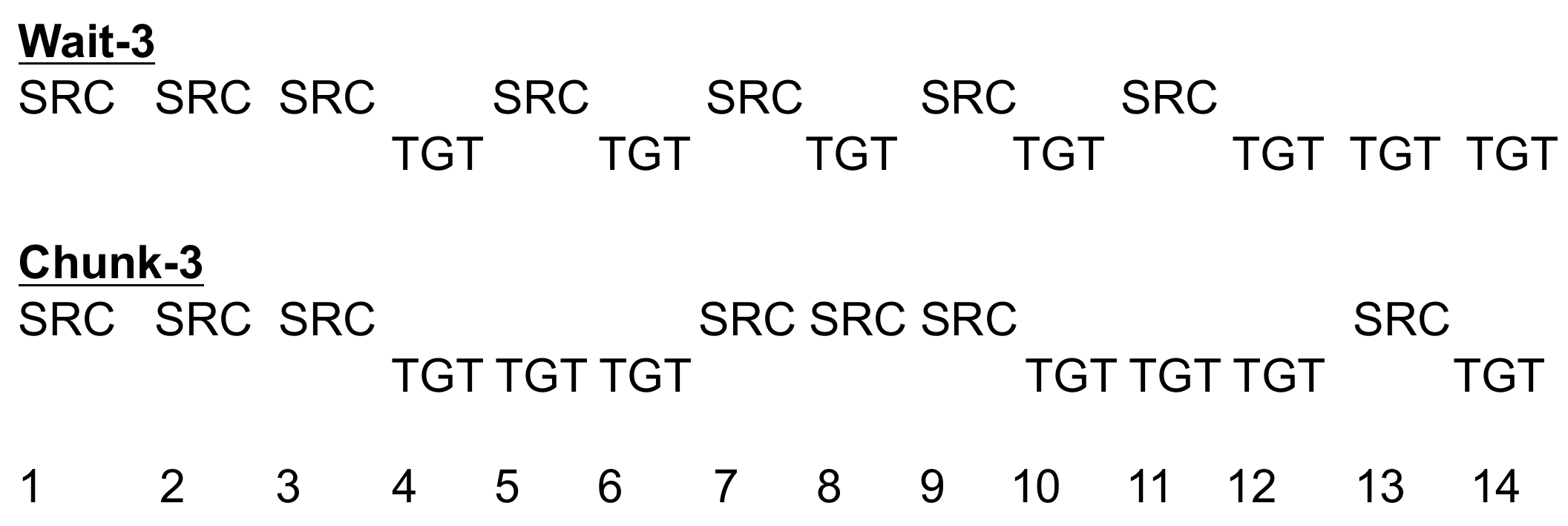}}
\caption{wait-3 and chunk-3 in a step-wise view.}
\label{fig:current_timing}
\vspace{5mm}
\centering
\centerline{\includegraphics[width=6.5cm]{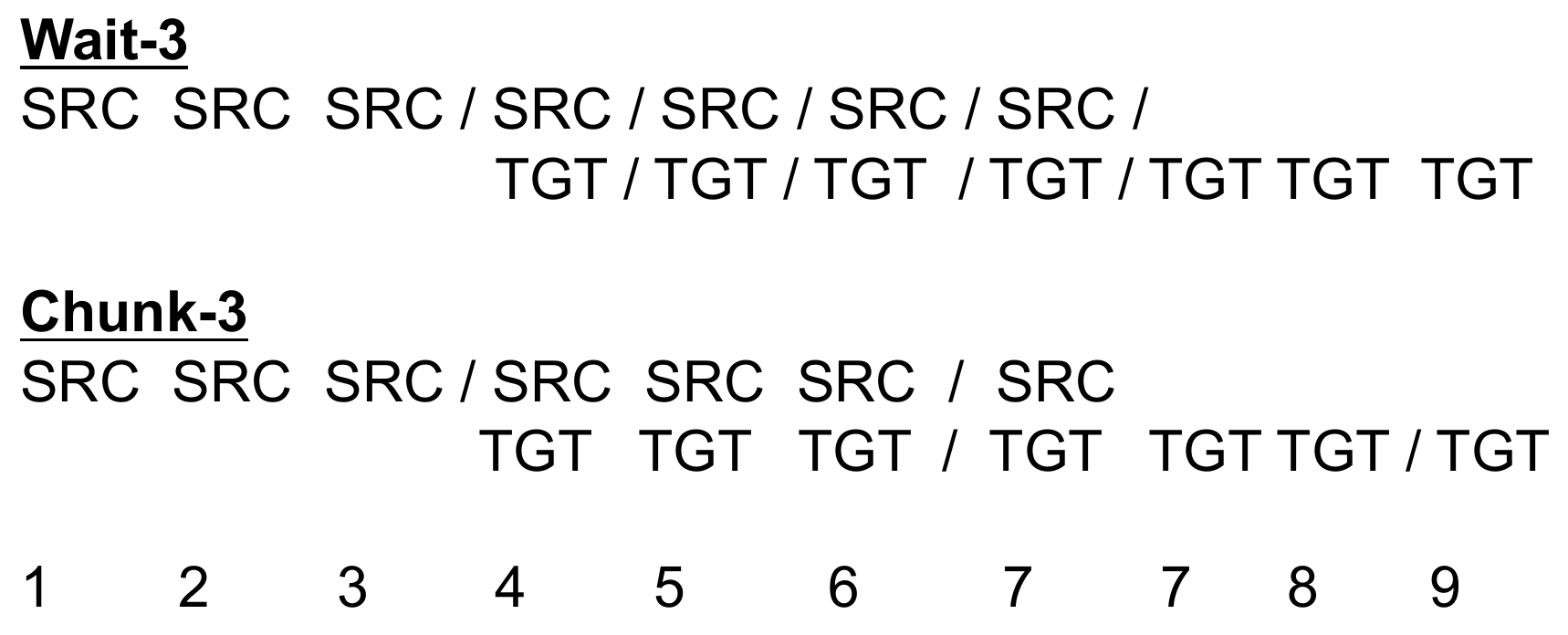}}
\caption{wait-3 and chunk-3 in a time-synchronous view.}
\label{fig:ideal_timing}
\end{figure}

This work focuses on a problem of AL.
AL does not sufficiently consider the cases where chunk-level outputs are generated at a time and give smaller latency values for them than one-by-one cases like wait-$k$.

Suppose we have a pair of seven input tokens and seven output tokens and apply two different simultaneous policies to this pair as shown in \autoref{fig:current_timing}:
wait-$3$ and \emph{chunk-3} that writes three tokens after reading three tokens.
\autoref{fig:ideal_timing} illustrates a situation where one input or output token requires one unit time to speak, ignoring computation time for simplicity.
These policies are equivalent to each other, from the viewpoint of latency in this situation.
However, AL for wait-$3$ is $\frac{15}{5} = 3$ and that for chunk-$3$ is $\frac{13}{7} \approx 1.857$.
AL tends to give a small latency value for such a long chunk-level output.
However, a long output delays the start of the translation of later parts and makes the listener feel the delay.
This problem becomes more severe when we need speech outputs because the speech outputs must be sequential.
Therefore, the length of a translation output has a large effect on delay and should be included in the latency measurement.
This observation suggests the need of another latency metric to cope with such situations.

In this paper, we propose a novel latency metric called \emph{Average Token Delay} (ATD) \footnote{ATD will be implemented in \url{https://github.com/facebookresearch/SimulEval}} that focuses on the end timings of partial translations.
ATD generalizes the latency measurement for simultaneous translation both with speech and text outputs and works intuitively for chunk-based outputs that are not properly handled by AL as presented above.
We present some simulation results to show the characteristics of ATD clearly and also demonstrate its effectiveness through simultaneous translation experiments.
\section{Simultaneous Machine Translation}

We quickly review the formulation of simultaneous translation to share the notation used in this paper.

In standard sentence-level NMT,
letting $\bm{x} = x_1,x_2,...,x_m$ be an input sentence and $\bm{y} = y_1,y_2,...,y_n$ be its translation,
the probability of the output is denoted as:
\begin{equation}
p(\bm{y}|\bm{x}) = \prod_{\tau=1}^{n} P(y_\tau | \bm{x}, \bm{y}_{<\tau}).
\end{equation}

Simultaneous translation takes a prefix of the input for its incremental decoding, formulated as follows:
\begin{equation}
p(\bm{y}|\bm{x}) = \prod_{\tau=1}^{n} P(y_\tau | \bm{x}_{\leq g(\tau)},\bm{y}_{<\tau}),
\end{equation}
where $g(\tau)$ is a monotonic non-decreasing function that represents the number of input tokens read until the prediction of $\tau$-th output so that $\bm{x}_{\leq g(\tau)}$ means an input prefix, and $\bm{y}_{<\tau}$ is a prefix translation predicted so far.
This means that we can obtain a pair of a input prefix and the corresponding prefix translation $(\bm{x}_{\leq g(\tau)}, \bm{y}_{\leq \tau})$ by that time.

The incremental decoding can be represented by a sequence of actions, \textit{READ} and \textit{WRITE}.
\textit{READ} is the action to take a new input, typically one token for text input or fixed number of frames for speech input.
\textit{WRITE} is the action to predict an output, typically one token for text output or the corresponding speech signal for speech output.
In this paper, we use $t$ ($0 \leq t \leq |\bm{x}|+|\bm{y}|$) for an index in such an action sequence and $\tau$ ($0 \leq \tau \leq |\bm{y}|$) for an index in the output.
\section{Existing Latency Metrics for Simultaneous Translation}

\subsection{Consecutive Wait (CW)}
\citet{gu-etal-2017-learning} proposed a latency metric called Consecutive Wait (CW) to measure the local delay.
CW at the step $t$ is denoted as follows:
$$
\mathrm{CW}(t)= 
\begin{cases}
\mathrm{CW}(t-1)+1 & a_{t}=\text { \textit{READ} } \\
0 & a_{t}=\text { \textit{WRITE} }
\end{cases}
$$
where $\mathrm{CW}(0)=0$, $a_{t}$ means an action at the step $t$.

Since CW is not originally an average measure of the latency, \citet{ma-etal-2019-stacl} used a step-wise CW: $\mathrm{CW}_{g}(\tau)=g(\tau)-g(\tau -1)$ for a policy $g(\cdot)$ to calculate an average CW for a sentence pair $(\bm{x}, \bm{y})$ as follows:

\begin{small}
\begin{equation}
\mathrm{CW}_{g}(\bm{x}, \bm{y})=\frac{\sum_{\tau=1}^{|\bm{y}|} \mathrm{CW}_{g}(\tau)}{\sum_{\tau=1}^{|\bm{y}|} \mathbbm{1}_{\mathrm{CW}_{g}(\tau)>0}}=\frac{|\bm{x}|}{\sum_{\tau=1}^{|\bm{y}|} \mathbbm{1}_{\mathrm{CW}_{g}(\tau)>0}}
\end{equation}
\end{small}

$g(\tau)$ is a monotonic non-decreasing function which represents number of source words read to predict the $\tau$-th target word.
This average CW becomes larger when \textit{WRITE} actions happen consecutively because $\mathrm{CW}_{g}(\tau) = 0$ when $g(\tau) - g(\tau -1) = 0$, i.e., the $\tau$-th and $(\tau -1)$-th output tokens are generated in the consecutive steps.
It is not intuitively appropriate to measure the latency of a sentence pair.
The average CW result in the same value for the following two extreme cases:
(1) all the output tokens are generated after reading all input tokens ($a_{1} = ... = a_{|\bm{x}|} = \text{\textit{READ}}$, $a_{|\bm{x}|+1} = ... = a_{|\bm{x}|+|\bm{y}|} = \text{\textit{WRITE}}$)
and (2) all the output tokens are generated after reading only the first input token ($a_{1} = \text{\textit{READ}}$, $a_{2} = ... = a_{|\bm{y}|+1} = \text{\textit{WRITE}}$, $a_{|\bm{y}|+2} = ... = a_{|\bm{x}|+|\bm{y}|} = \text{\textit{READ}}$).

\subsection{Average Propotion (AP)}
\citet{cho-2016} proposed Average Proportion (AP).
AP for a sentence pair $(\bm{x}, \bm{y})$ is denoted as follows:
\begin{equation}
\mathrm{AP}_{g}(\bm{x}, \bm{y})=\frac{1}{|\bm{x}| |\bm{y}|} \sum_{\tau} g(\tau)
\end{equation}

Suppose we use the wait-1 policy.
The sum $\sum_{\tau} g(\tau)$ is given by $\sum_{\tau=1}^{|\bm{y}|} \tau$.
For a sentence pair with $|\bm{x}_1| = |\bm{y}_1| = 1$, $\mathrm{AP}(\bm{x}_1, \bm{y}_1) = 1$.
When the length becomes longer, e.g., $|\bm{x}_{10}| = |\bm{y}_{10}| = 10$, $\mathrm{AP}(\bm{x}_1, \bm{y}_1) = 0.55$.
AP differs with the change of the sequence length for the same simultaneous translation policy and is not intuitive for the latency metric.

\subsection{Average Lagging (AL)}
\citet{ma-etal-2019-stacl} proposed Average Lagging (AL).
AL is denoted as follows:
\begin{equation}
AL_g(\bm{x}, \bm{y}) = \frac{1}{\tau_g(|\bm{x}|)} \sum_{\tau =1}^{\tau_g(|\bm{x}|)} \left( g(\tau) - \frac{\tau -1}{r} \right).
\label{eqn:AL_g}
\end{equation}
where r is the length ratio defined as $|\bm{y}|/|\bm{x}|$ and $\tau_g(|\bm{x}|)$ is the cut-off step defined as follows:
\begin{equation}
\tau_g(|\bm{x}|) = \min\{\tau \mid g(\tau) = |\bm{x}|\}
\label{eqn:tau_g}
\end{equation}
meaning the index of the output token predicted right after the observation of the entire source sentence.

AL solves the problem of AP mentioned above by focusing on the difference from the \emph{ideal} policy based on the length ratio.
However, AL still suffers from unintuitive latency measurement because AL can be negative when the model finishes the translation before reading the entire input due to the subtraction term.
\citet{ma-etal-2020-simuleval} modified AL by changing the calculation of the length ratio $r$ based on the length of the reference translation.
\citet{papi-etal-2022-generation} proposed Length-Adaptive Average Lagging (LAAL) that uses the longer one between the reference and output.
These modifications do not completely avoid negative values when the model generates a long partial output faster than the ideal policy.
\cite{arivazhagan-etal-2019-monotonic} proposed another variant, Differentiable Average Lagging (DAL), for optimizing simultaneous translation model.
\citet{iranzo-sanchez-etal-2021-stream-level} proposed a method to calculate AL for a streaming input in a segmentation-free manner.

\section{Proposed Metric: Average Token Delay}

We propose a novel latency metric called Average Token Delay (ATD).
We start from the latency measurement using ATD in case of simultaneous speech-to-speech MT and then generalize it for speech-to-text and text-to-text cases.

\begin{figure}[t]
\centering
\centerline{\includegraphics[width=8.5cm]{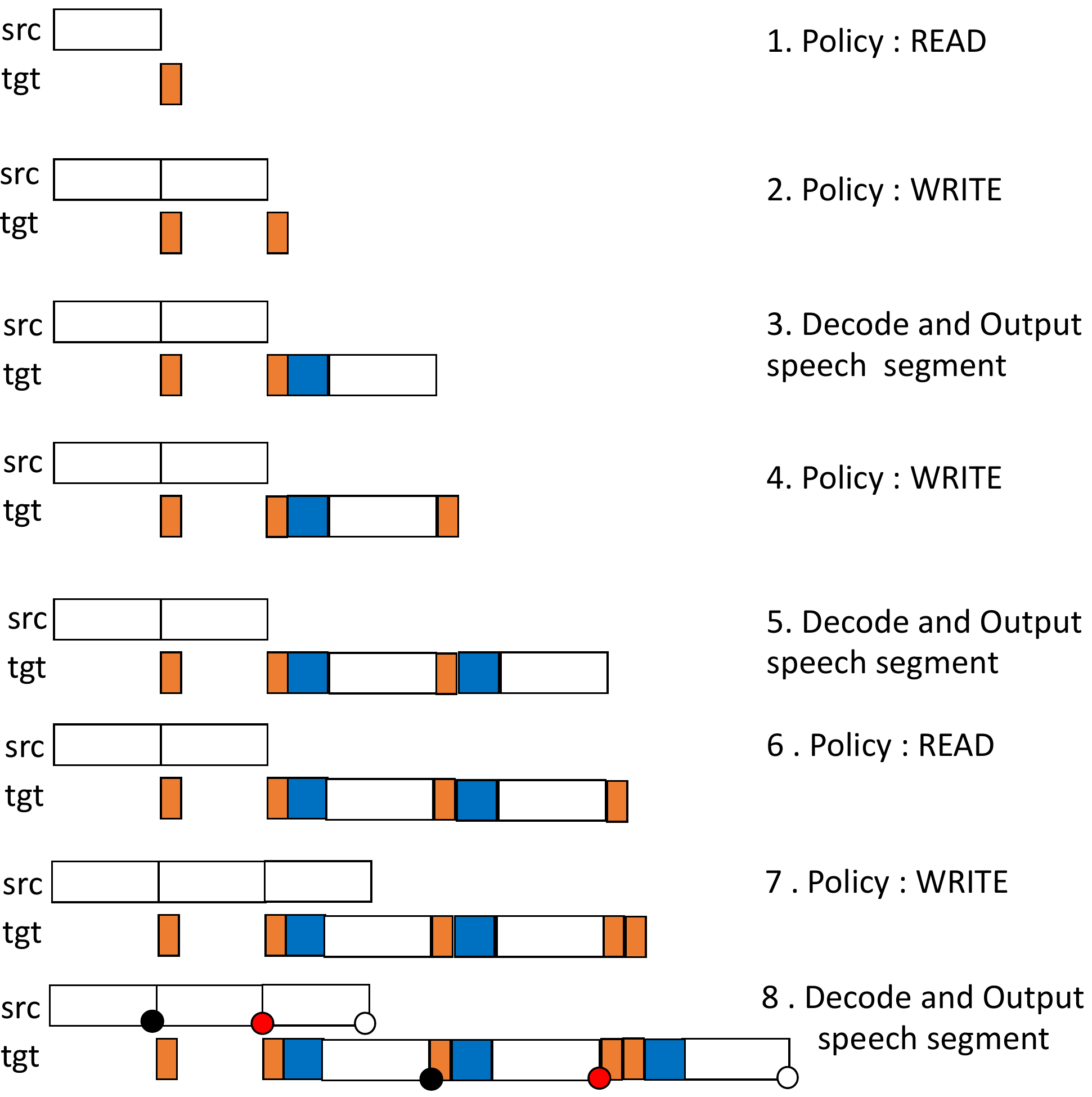}}
\caption{Step-by-step example of simultaneous speech-to-speech MT}
\label{fig:s2s_simulate}
\end{figure}

\subsection{ATD for simultaneous speech-to-speech translation}
\autoref{fig:s2s_simulate} illustrates a step-by-step workflow of simultaneous speech-to-speech MT.
In this figure, the white boxes represent fixed-length speech segments, the orange ones represent the processing time to encode prefix inputs and to judge whether prefix translations can be predicted there, and the blue ones represent decoding time.
Intuitively, ATD is the average time difference between the points of the same color (black, red, and white). 

Suppose we have an input $\bm{x}$ and an output $\bm{y}$ segmented into $C$ chunks $\bm{x} = \bm{x}^{1}, \bm{x}^{2}, ..., \bm{x}^{C}$ and $\bm{y} = \bm{y}^{1}, \bm{y}^{2}, ..., \bm{y}^{C}$, respectively.
An input chunk $\bm{x}^{c}$ is what is observed after the previous judgment for the prefix translation $\bm{y}^{c-1}$ and is used to predict $\bm{y}^{c}$.
In the case of text, each chunk consists of sub-segments with words, subwords, or characters.
In the case of speech, we divide each chunk into sub-segments of 0.3 seconds long from the beginning of the chunk, assuming one word is uttered in 0.3 seconds.
Through this segmentation, the input and output sentences can be represented as a series of sub-segments $\bm{x} = x_1, ..., x_{|\bm{x}|}$, $\bm{y} = y_1, ..., y_{|\bm{y}|}$ respectively.

ATD is defined as follows:
\begin{equation}
\textrm{ATD}(\bm{x}, \bm{y})= \frac{1}{|\bm{y}|}\sum_{t=1}^{|\bm{y}|} \left( T(y_{t}) - T(x_{a(t)}) \right)
\label{eqn:atd_definition}
\end{equation}
where
\begin{equation}
\label{s_condition}
a(t) = \begin{cases}s(t) & \textrm{if}\ s(t) \leq L_{acc}(\bm{x}^{c(t)}) \\
L_{acc}( \bm{x}^{c(t)}) & \textrm{otherwise}\end{cases}
\end{equation}
\begin{equation}
\label{max_condition}
s(t) = t - \max(L_{acc}(\bm{y}^{c(t) -1}) - L_{acc}(\bm{x}^{c(t) -1}),0)
\end{equation}

$T(\cdot)$ in Eq.~(\ref{eqn:atd_definition}) represents the ending time of each token, which is shown as colored points in \autoref{fig:s2s_simulate}.
$a(t)$ represents the index of the input token corresponding to $y_{t}$.
$L_{acc} (\bm{x}^{c}) = \sum_{j =1}^{c} |\bm{x}^{j}| $ is the cumulative length up to the $c$-th chunk, and  $L_{acc} (\bm{x}^{0}) = 0$. 
$L_{acc} (\bm{y}^{c})$ is defined similarly.
$c(t)$ denotes the chunk number $c$ to which $y_t$ belongs.
As shown in Eq.~(\ref{max_condition}), if the previous translation prefix is longer than the previous input prefix, $s(t)$ becomes smaller than the output index $t$, which means the previous long output makes the time difference between the input token and the corresponding output token larger.

ATD is the average delay of output sub-segments against their corresponding input sub-segments, considering the latency required for inputs and outputs.
Although the input-output correspondence does not necessarily mean semantic equivalence, especially for language pairs with large differences in their word order and the numbers of tokens, we use this simplified formulation for the latency measurement as same as AL.

\begin{figure}[t]
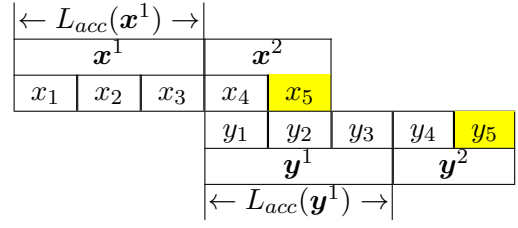
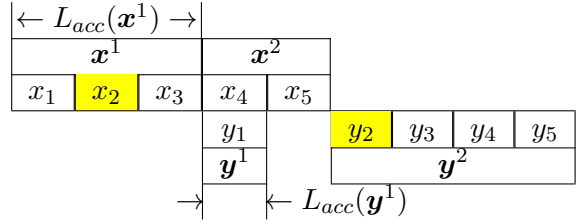
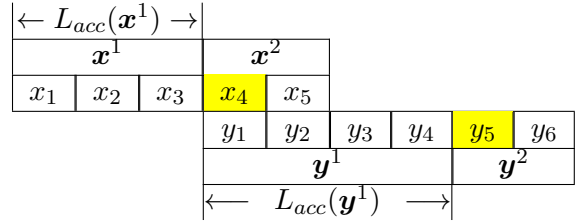

\centering

\begin{minipage}{1.0\hsize}
\centering
\begin{tabular}{cccccccc}
\multicolumn{3}{|c|}{ $\hspace{-3mm}\leftarrow L_{acc} ( \bm{x}^{1}) \rightarrow \hspace{-3mm}$} & & & & & \\
\cline{1-5}
\multicolumn{3}{|c|}{$\bm{x}^1$} & \multicolumn{2}{|c|}{$\bm{x}^2$} & & & \\
\cline{1-5}
\multicolumn{1}{|c|}{$x_1$} & \multicolumn{1}{|c|}{$x_2$} & \multicolumn{1}{|c|}{$x_3$} & \multicolumn{1}{|c|}{$x_4$} & \multicolumn{1}{|c|}{\cellcolor[rgb]{1.0, 1.0, 0.0}$x_5$} & & & \\ \hline
& & & \multicolumn{1}{|c|}{$y_1$} & \multicolumn{1}{|c|}{$y_2$} & \multicolumn{1}{|c|}{$y_3$} & \multicolumn{1}{|c|}{$y_4$} & \multicolumn{1}{|c|}{\cellcolor[rgb]{1.0, 1.0, 0.0}$y_5$} \\ \cline{4-8}
& & & \multicolumn{3}{|c|}{$\bm{y}^1$} & \multicolumn{2}{|c|}{$\bm{y}^2$} \\ \cline{4-8}
& & & \multicolumn{3}{|c|}{ $\hspace{-3mm}\leftarrow L_{acc} ( \bm{y}^{1}) \rightarrow \hspace{-3mm}$ } & &
\end{tabular}
\subcaption{$L_{acc}(\bm{y}^{c(t) -1}) - L_{acc} ( \bm{x}^{c(t) -1}) = 0$ ($t=5$)}
\label{tab:equation1a}
\end{minipage}

\vspace{3mm}

\begin{minipage}{1.0\hsize}
\centering
\begin{tabular}{ccccccccc}
\multicolumn{3}{|c|}{ $\hspace{-3mm}\leftarrow L_{acc} ( \bm{x}^{1}) \rightarrow \hspace{-3mm}$} & & & & & &\\
\cline{1-5}
\multicolumn{3}{|c|}{$\bm{x}^1$} & \multicolumn{2}{|c|}{$\bm{x}^2$} & & & & \\
\cline{1-5}
\multicolumn{1}{|c|}{$x_1$} & \multicolumn{1}{|c|}{\cellcolor[rgb]{1.0, 1.0, 0.0}$x_2$} & \multicolumn{1}{|c|}{$x_3$} & \multicolumn{1}{|c|}{$x_4$} & \multicolumn{1}{|c|}{$x_5$} & & & & \\ \hline
& & & \multicolumn{1}{|c|}{$y_1$} & & \multicolumn{1}{|c|}{\cellcolor[rgb]{1.0, 1.0, 0.0}$y_2$} & \multicolumn{1}{|c|}{$y_3$} & \multicolumn{1}{|c|}{$y_4$} & \multicolumn{1}{|c|}{$y_5$} \\ \cline{4-4} \cline{6-9}
& & & \multicolumn{1}{|c|}{$\bm{y}^1$} & & \multicolumn{4}{|c|}{$\bm{y}^2$} \\ \cline{4-4} \cline{6-9}
& & \multicolumn{1}{r}{$\rightarrow \hspace{-2.5mm}$} & \multicolumn{1}{|c|}{ } & \multicolumn{4}{l}{$\hspace{-2.5mm} \leftarrow L_{acc} ( \bm{y}^{1})$} &
\end{tabular}
\subcaption{$L_{acc}(\bm{y}^{c(t) -1}) - L_{acc} ( \bm{x}^{c(t) -1}) < 0$ ($t=2$)}
\label{tab:equation1c}
\end{minipage}

\vspace{3mm}

\begin{minipage}{1.0\hsize}
\centering
\begin{tabular}{ccccccccc}
\multicolumn{3}{|c|}{ $\hspace{-3mm}\leftarrow L_{acc} ( \bm{x}^{1}) \rightarrow \hspace{-3mm}$} & & & & & &\\
\cline{1-5}
\multicolumn{3}{|c|}{$\bm{x}^1$} & \multicolumn{2}{|c|}{$\bm{x}^2$} & & & & \\
\cline{1-5}
\multicolumn{1}{|c|}{$x_1$} & \multicolumn{1}{|c|}{$x_2$} & \multicolumn{1}{|c|}{$x_3$} & \multicolumn{1}{|c|}{\cellcolor[rgb]{1.0, 1.0, 0.0}$x_4$} & \multicolumn{1}{|c|}{$x_5$} & & & & \\ \hline
& & & \multicolumn{1}{|c|}{$y_1$} & \multicolumn{1}{|c|}{$y_2$} & \multicolumn{1}{|c|}{$y_3$} & \multicolumn{1}{|c|}{$y_4$} & \multicolumn{1}{|c|}{\cellcolor[rgb]{1.0, 1.0, 0.0}$y_5$} & \multicolumn{1}{|c|}{$y_6$} \\ \cline{4-9}
& & & \multicolumn{4}{|c|}{$\bm{y}^1$} & \multicolumn{2}{|c|}{$\bm{y}^2$} \\ \cline{4-9}
& & & \multicolumn{4}{|c|}{ $\hspace{-3mm}\longleftarrow \hspace{2mm} L_{acc} ( \bm{y}^{1}) \hspace{2mm} \longrightarrow \hspace{-3mm}$ } & & 
\end{tabular}
\subcaption{$L_{acc}(\bm{y}^{c(t) -1}) - L_{acc} ( \bm{x}^{c(t) -1}) > 0$ ($t=5$)}
\label{tab:equation1b}
\end{minipage}

\caption{Examples for the explanation of Eq.~(\ref{max_condition})}
\label{tab:equation1}
\end{figure}

\begin{figure}[t]
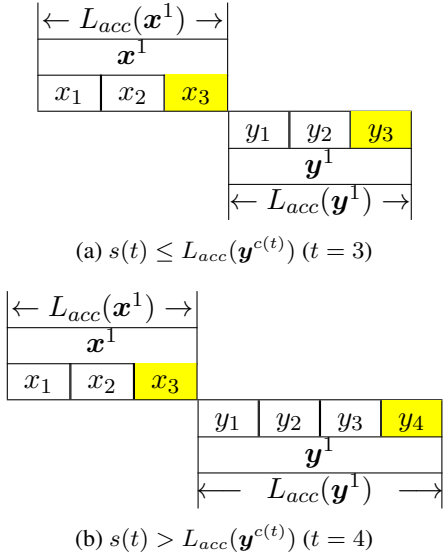

\centering

\begin{minipage}{1.0\hsize}
\centering
\begin{tabular}{cccccc}
\multicolumn{3}{|c|}{ $\hspace{-3mm}\leftarrow L_{acc} ( \bm{x}^{1}) \rightarrow \hspace{-3mm}$} & & & \\
\cline{1-3}
\multicolumn{3}{|c|}{$\bm{x}^1$} & & & \\
\cline{1-3}
\multicolumn{1}{|c|}{$x_1$} & \multicolumn{1}{|c|}{$x_2$} & \multicolumn{1}{|c|}{\cellcolor[rgb]{1.0, 1.0, 0.0}{$x_3$}} & & & \\ \hline
& & & \multicolumn{1}{|c|}{$y_1$} & \multicolumn{1}{|c|}{$y_2$} & \multicolumn{1}{|c|}{\cellcolor[rgb]{1.0, 1.0, 0.0}$y_3$} \\ \cline{4-6}
& & & \multicolumn{3}{|c|}{$\bm{y}^1$} \\ \cline{4-6}
& & & \multicolumn{3}{|c|}{ $\hspace{-3mm}\leftarrow L_{acc} ( \bm{y}^{1}) \rightarrow \hspace{-3mm}$ }
\end{tabular}
\subcaption{$s(t) \leq L_{acc}(\bm{y}^{c(t)})$ ($t=3$)}
\label{tab:equation2a}
\end{minipage}

\vspace{3mm}

\begin{minipage}{1.0\hsize}
\centering
\begin{tabular}{ccccccc}
\multicolumn{3}{|c|}{ $\hspace{-3mm}\leftarrow L_{acc} ( \bm{x}^{1}) \rightarrow \hspace{-3mm}$} & & & & \\
\cline{1-3}
\multicolumn{3}{|c|}{$\bm{x}^1$} & & & & \\
\cline{1-3}
\multicolumn{1}{|c|}{$x_1$} & \multicolumn{1}{|c|}{$x_2$} & \multicolumn{1}{|c|}{\cellcolor[rgb]{1.0, 1.0, 0.0}$x_3$} & & & & \\ \hline
& & & \multicolumn{1}{|c|}{$y_1$} & \multicolumn{1}{|c|}{$y_2$} & \multicolumn{1}{|c|}{$y_3$} & \multicolumn{1}{|c|}{\cellcolor[rgb]{1.0, 1.0, 0.0}$y_4$} \\ \cline{4-7}
& & & \multicolumn{4}{|c|}{$\bm{y}^1$} \\ \cline{4-7}
& & & \multicolumn{4}{|c|}{ $\hspace{-3mm}\longleftarrow \hspace{2mm} L_{acc} ( \bm{y}^{1}) \hspace{2mm} \longrightarrow \hspace{-3mm}$ }
\end{tabular}
\subcaption{$s(t) > L_{acc}(\bm{y}^{c(t)})$ ($t=4$)}
\label{tab:equation2b}
\end{minipage}
\caption{Example for the explanation of Eq.~(\ref{s_condition})}
\label{tab:equation2}
\end{figure}

\autoref{tab:equation1} shows examples to explain the term in $\max$ operator in Eq.~(\ref{max_condition}).
In \autoref{tab:equation1a}, Suppose we measure the token delay on $y_5$.
$\bm{y}_5$ is in the second output chunk, so $c(5) = 2$.
Since $L_{acc}(\bm{y}^{1}) = L_{acc} ( \bm{x}^{1}) = 3$, we obtain $a(5) = s(5) = 5 - 0 = 5 \leq L_{acc}(\bm{x}^{2}) = 5$, and therefore $\bm{y}_5$ corresponds to $\bm{x}_5$.
In \autoref{tab:equation1c}, Suppose we measure the token delay on $y_2$.
$y_2$ is in the second output chunk, so c(2) = 2.
Since $L_{acc}(\bm{y}^{1}) - L_{acc}(\bm{x}^{1}) = 1 - 3  < 0 $, we obtain $a(2) = s(2) = 2 - 0 = 2 \leq L_{acc}(x^2) = 5$, therefore $y_2$ corresponds to $x_2$.
In \autoref{tab:equation1b}, the first output chunk is longer: $L_{acc}(\bm{y}^{1}) = 4$.
This results in $a(5) = s(5) = 5 - 1 = 4 \leq L_{acc}(\bm{x}^{2}) = 5$, so $\bm{y}_5$ corresponds to $\bm{x}_4$.

\autoref{tab:equation2} shows examples to explain Eq.~(\ref{s_condition}).
In \autoref{tab:equation2a}, we measure the token delay on $y_3$.
Here, we obtain $s(t) = 3 \leq L_{acc}( \bm{x}^{1}) = 3$, so $y_3$ corresponds to $x_3$.
\autoref{tab:equation2b}, we measure the token delay on $y_4$.
Here, we obtain $s(t) = 4 > L_{acc}( \bm{x}^{1}) = 3$, so $y_4$ corresponds to $x_3$.

\begin{figure}[t]
\centering
\centerline{\includegraphics[height=2.0cm,width=6.0cm]{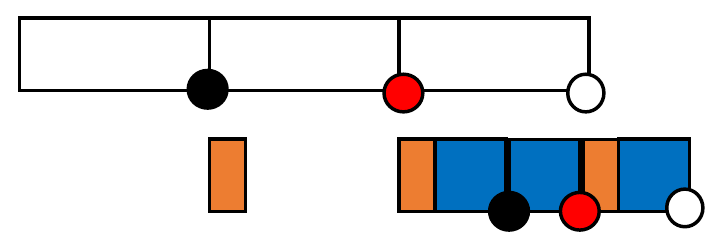}}
\caption{Summary view for latency measurement for simultaneous speech-to-text translation}
\label{fig:s2t_simulate}

\centering
\centerline{\includegraphics[height=2.0cm,width=4.3cm]{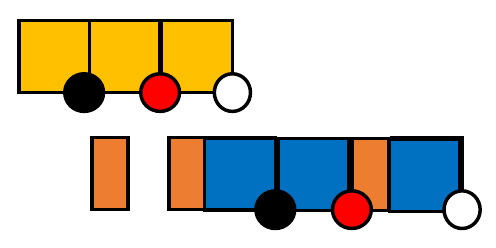}}
\caption{Summary view for latency measurement for simultaneous text-to-text translation}
\label{fig:t2t_simulate}

\centering
\centerline{\includegraphics[height=2.5cm,width=3.2cm]{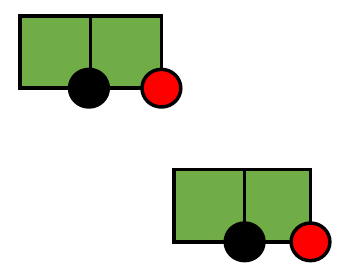}}
\caption{Summary view for non-computation-aware latency measurement for simultaneous text-to-text translation}
\label{fig:t2tNCA_simulate}
\end{figure}

\subsection{ATD for simultaneous \{speech,text\}-to-text translation}
\autoref{fig:s2t_simulate} illustrates the latency measurement for speech-to-text simultaneous translation, where the output duration can be ignored.
\autoref{fig:t2t_simulate} is the one for text-to-text simultaneous translation.
We reserve input duration here because input for text-to-text simultaneous translation comes from speech via automatic speech recognition (ASR), in most cases.
The input duration reflects ASR computation time.

\subsection{Non-computation-aware ATD}
We sometimes use the latency measurement independent of the computation time for estimating ideal situations that are not influenced by the performance of computers and the efficiency of implementations.
In Figures~\ref{fig:s2s_simulate}, \ref{fig:s2t_simulate}, and \ref{fig:t2t_simulate}, we remove the orange, blue and yellow parts and only include the duration of speech segments to calculate delay.
However, this means all the terms in text-to-text translation become 0.
We follow the conventional step-wise latency measurement as CW and AP by letting each input and output word spend one step as shown in \autoref{fig:t2tNCA_simulate}.
Also, we assume the model can read the next input and output the partial translation in parallel as shown in \autoref{fig:ideal_timing}.

\section{Simulation}
Before presenting the latency measurement experiments using real data,
we show simulation results comparing AL and ATD in different conditions in simultaneous text-to-text translation.

\subsection{Case 1: 40-40}\label{subsec:case1}

\begin{figure}[t]
\begin{minipage}[b]{1.0\hsize}
\centering
\centerline{\includegraphics[width=8.5cm]{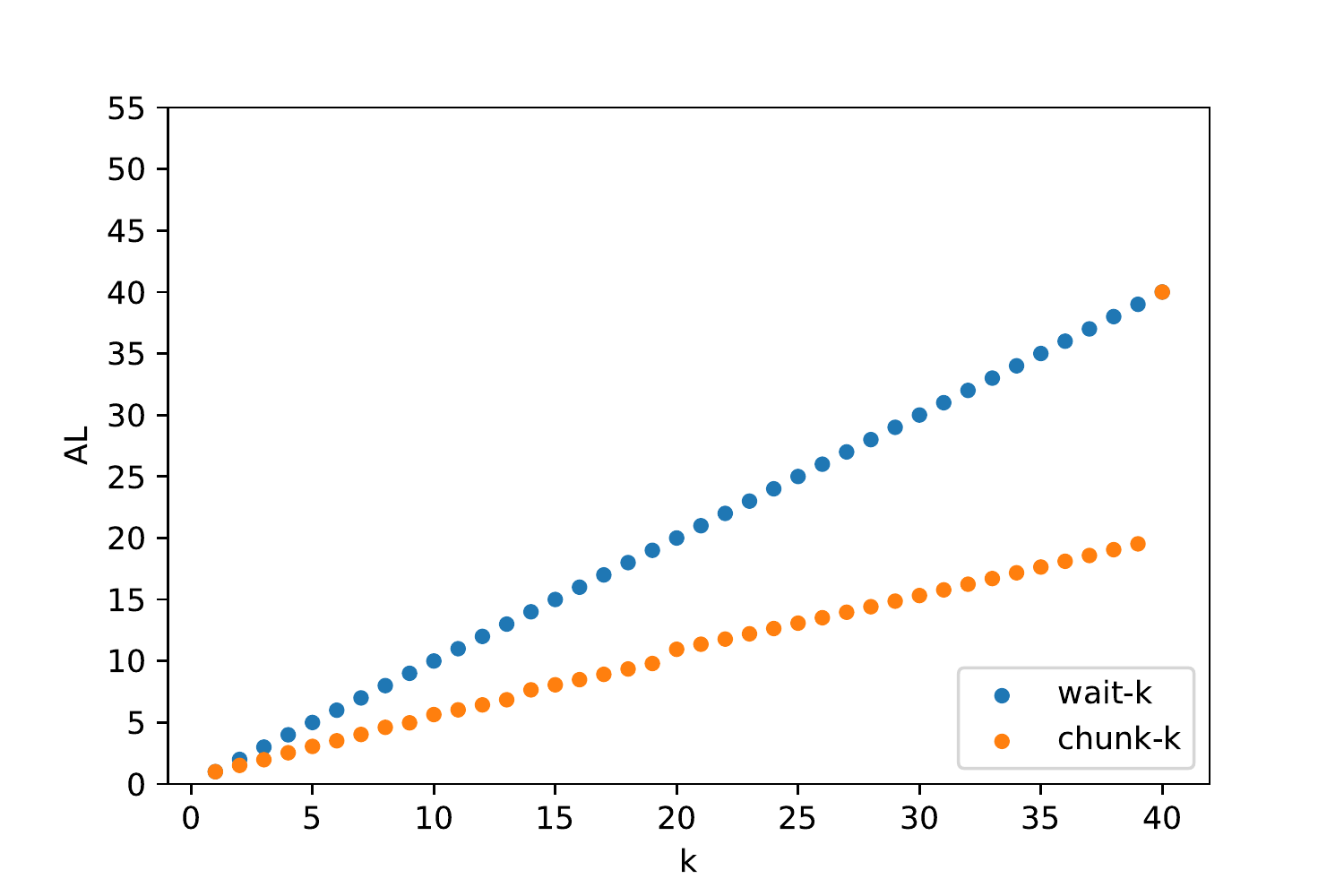}}
\subcaption{Latency measurement by AL (40-40)}\label{fig:ALsrc40tgt40}
\end{minipage}
\begin{minipage}[b]{1.0\hsize}
\centering
\centerline{\includegraphics[width=8.5cm]{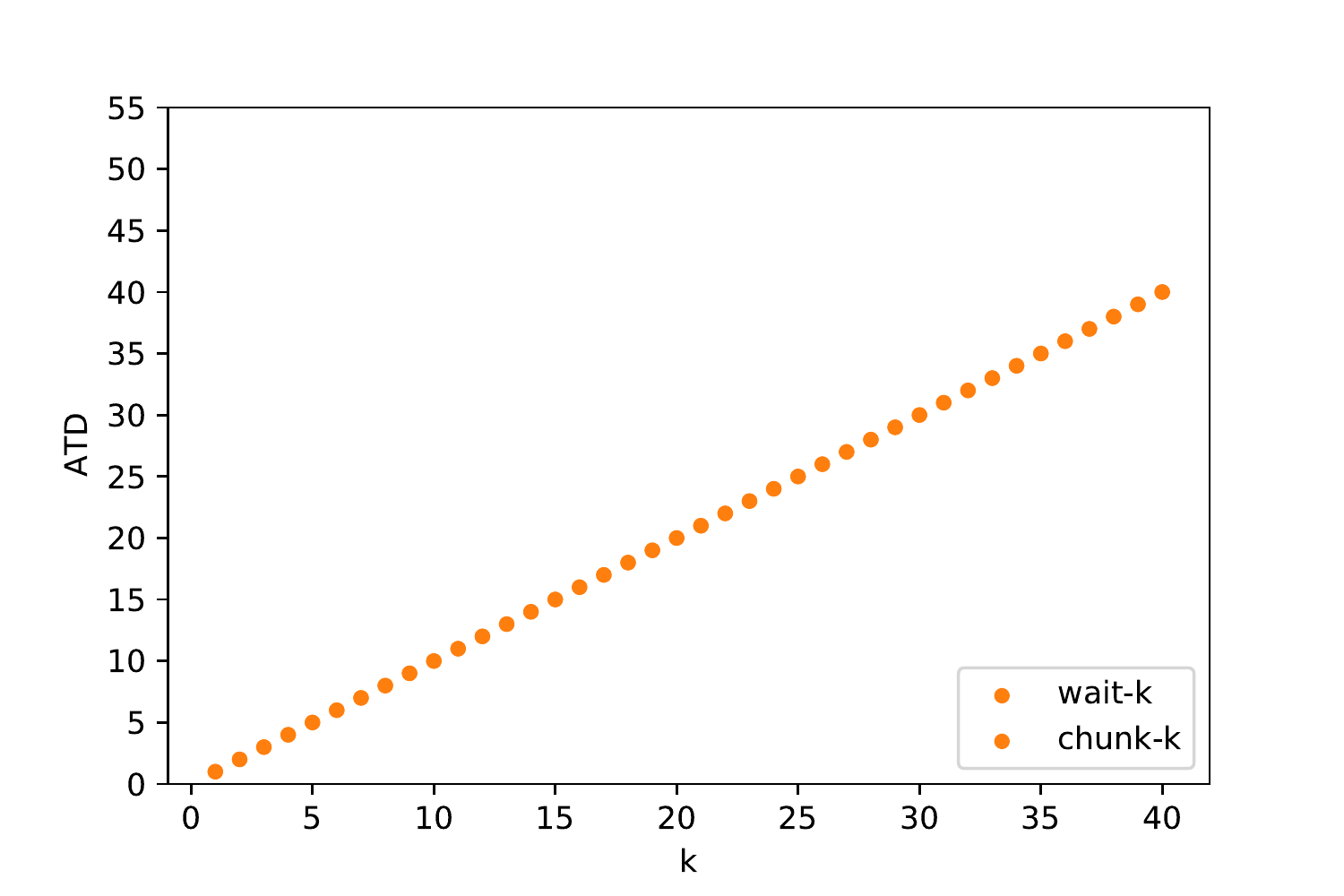}}
\subcaption{Latency measurement by ATD (40-40)}\label{fig:ATDsrc40tgt40}
\end{minipage}
\caption{Case 1 (40 input and 40 output tokens)}\label{fig:case1}
\end{figure}

We assume that the numbers of tokens of the input and output are both 40.
We compare the latency measurement by AL and ATD where the hyperparameter $k$ for wait-$k$ and chunk-$k$ varies from 1 to 40. 
Here, for simplicity, we assume the length of input and output chunks are the same for chunk-$k$ until the prediction of the end-of-sentence token.

\autoref{fig:case1} shows the results.
\autoref{fig:ALsrc40tgt40} indicates the gap between wait-$k$ and chunk-$k$ by AL mentioned in \autoref{sec:introduction}, while ATD results in the same values for them as shown in \autoref{fig:ATDsrc40tgt40}.
One serious problem raised here is the large jump in AL at $k=40$.
For example, in the case of chunk-$39$, AL uses $r=1$, $\tau_{g_{\textrm{chunk-}39}}(|\bm{x}|) = 40$, $g_{\textrm{chunk-}39}(\tau)=39$ ($1 \leq \tau \leq 39$), and $g_{\textrm{chunk-}39}(\tau)=40$ ($\tau = 40$).
Then the AL value becomes $\frac{1}{40} \left\{ \left( \sum_{\tau=1}^{39} \tau \right) + 1 \right\} = \frac{781}{40} = 19.525$.
However, in the case of chunk-$40$, $g_{\textrm{chunk-}40}(\tau) = 40$ for all $\tau$ and  $\tau_{g_{\textrm{chunk-}40}}(|\bm{x}|) = 1$ according to \autoref{eqn:tau_g}.
As a result, the AL value becomes 40.
This phenomenon comes from the definition of the cut-off step described in Section 4.2 by \citet{ma-etal-2019-stacl}, where they assume later outputs derive no further delays.

\subsection{Case 2: 40-100}\label{subsec:case2}

\begin{figure}[t]
\begin{minipage}[b]{1.0\hsize}
\centering
\centerline{\includegraphics[width=8.5cm]{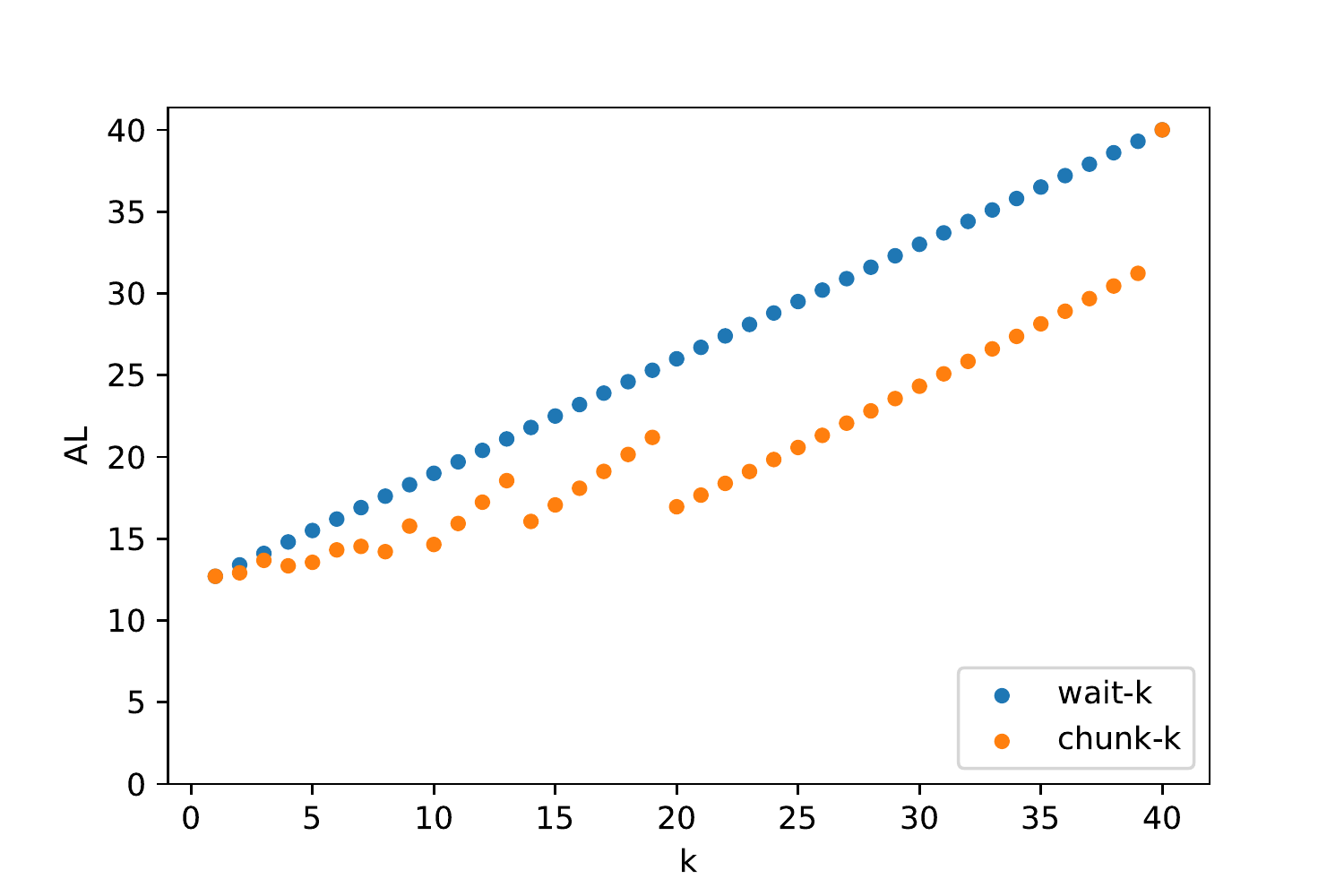}}
\subcaption{Latency measurement by AL (40-100)}\label{fig:ALsrc40tgt100}
\end{minipage}

\begin{minipage}[b]{1.0\hsize}
\centering
\centerline{\includegraphics[width=8.5cm]{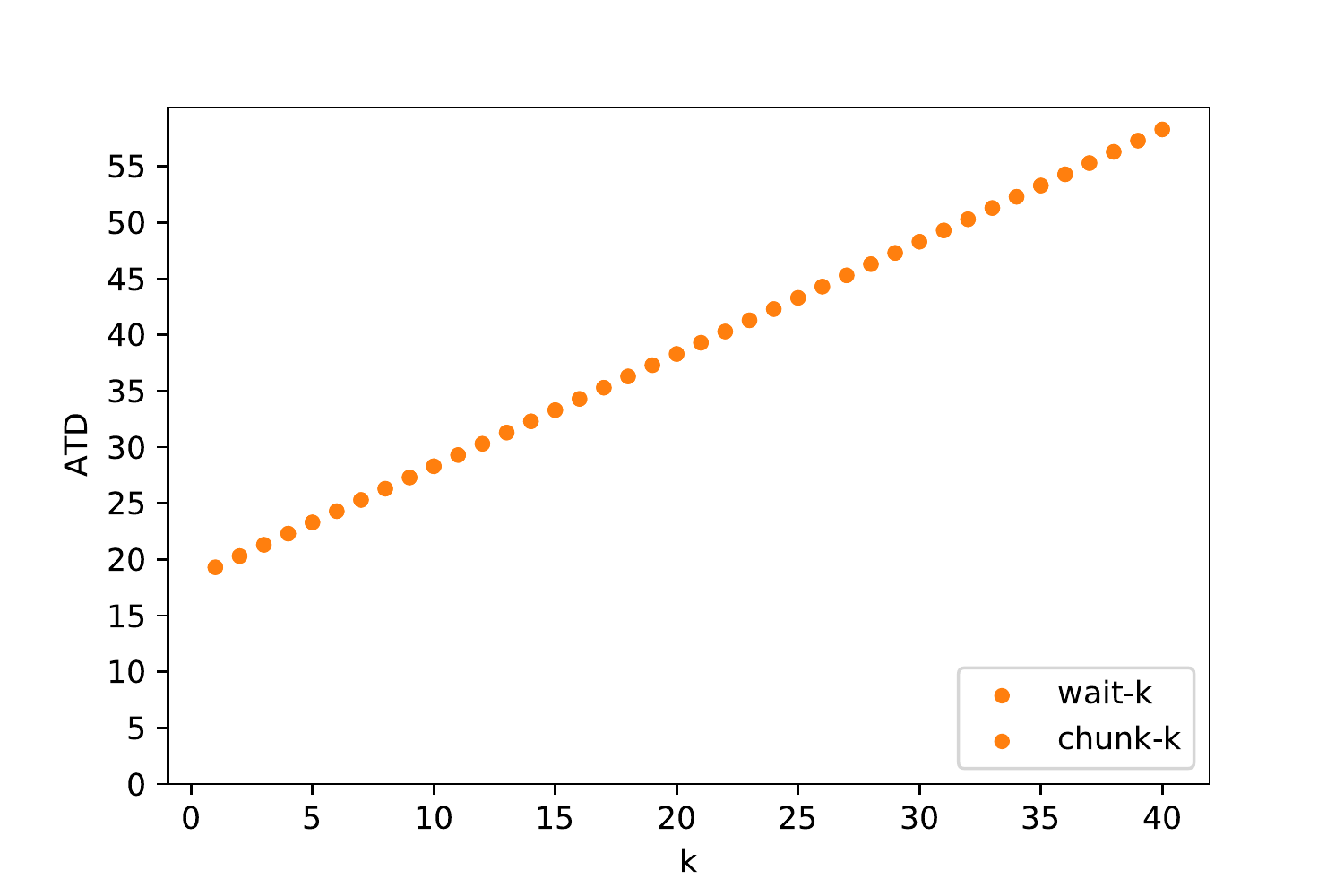}}
\subcaption{Latency measurement by ATD (40-100)}\label{fig:ATDsrc40tgt100}
\end{minipage}

\caption{Case 2 (40 input and 100 output tokens)}\label{fig:case2}
\end{figure}

Next, we simulate another situation with an input of 40 tokens and an output of 100 tokens and conduct the same analysis.
\autoref{fig:case2} shows the results.
As shown in \autoref{fig:ALsrc40tgt100}, AL suffers from discontinuous latency values for chunk-$k$ that are caused by similar phenomena in the example above.
In contrast, ATD results in the same linear correlations for wait-$k$ and chunk-$k$ as shown in \autoref{fig:ATDsrc40tgt100}.

\subsection{Case 3: 40-20}\label{subsec:case3}

\begin{figure}[t]
\begin{minipage}[b]{1.0\hsize}
\centering
\centerline{\includegraphics[width=8.5cm]{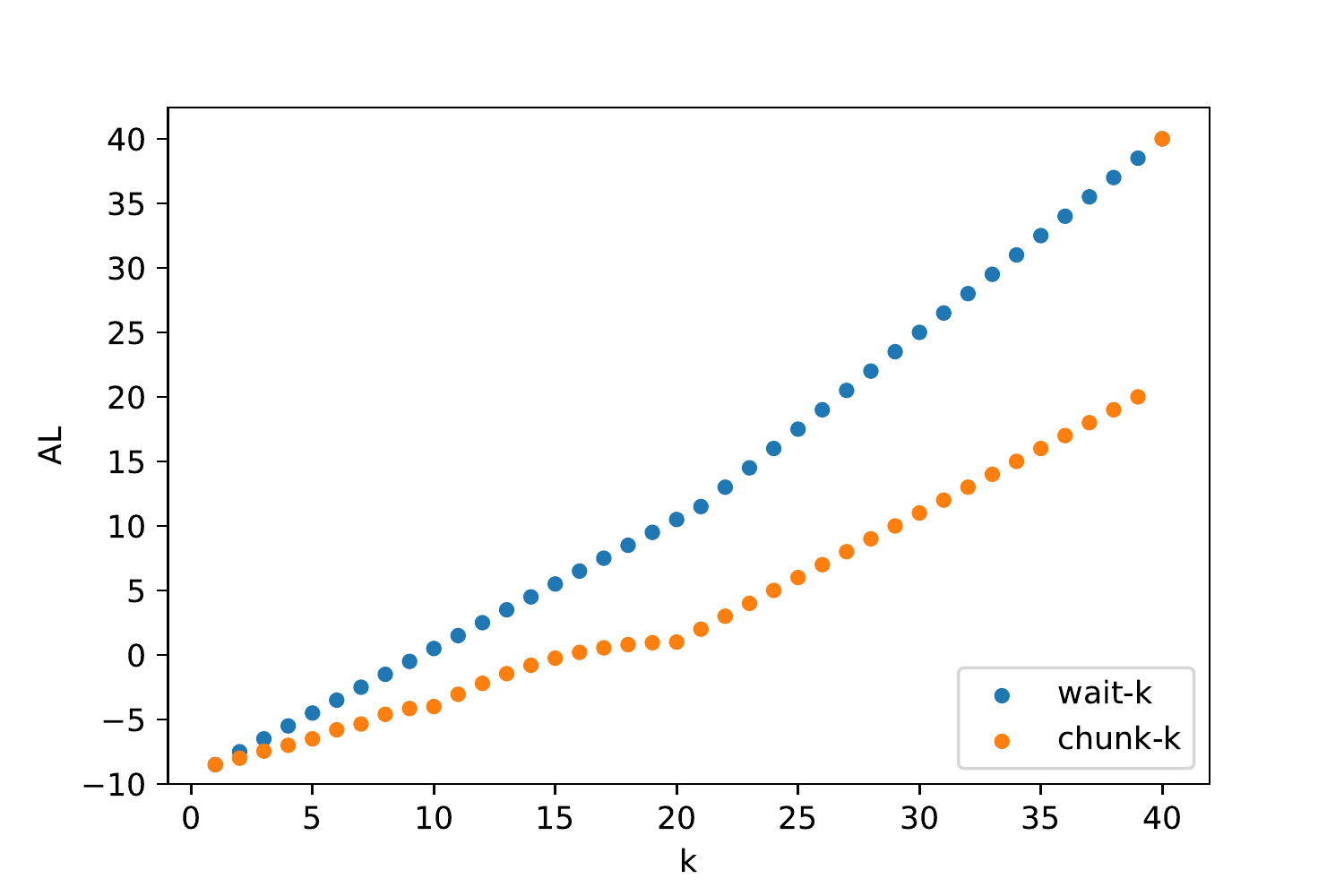}}
\subcaption{Latency measurement by AL (40-20)}\label{fig:ALsrc40tgt20}
\end{minipage}

\begin{minipage}[b]{1.0\hsize}
\centering
\centerline{\includegraphics[width=8.5cm]{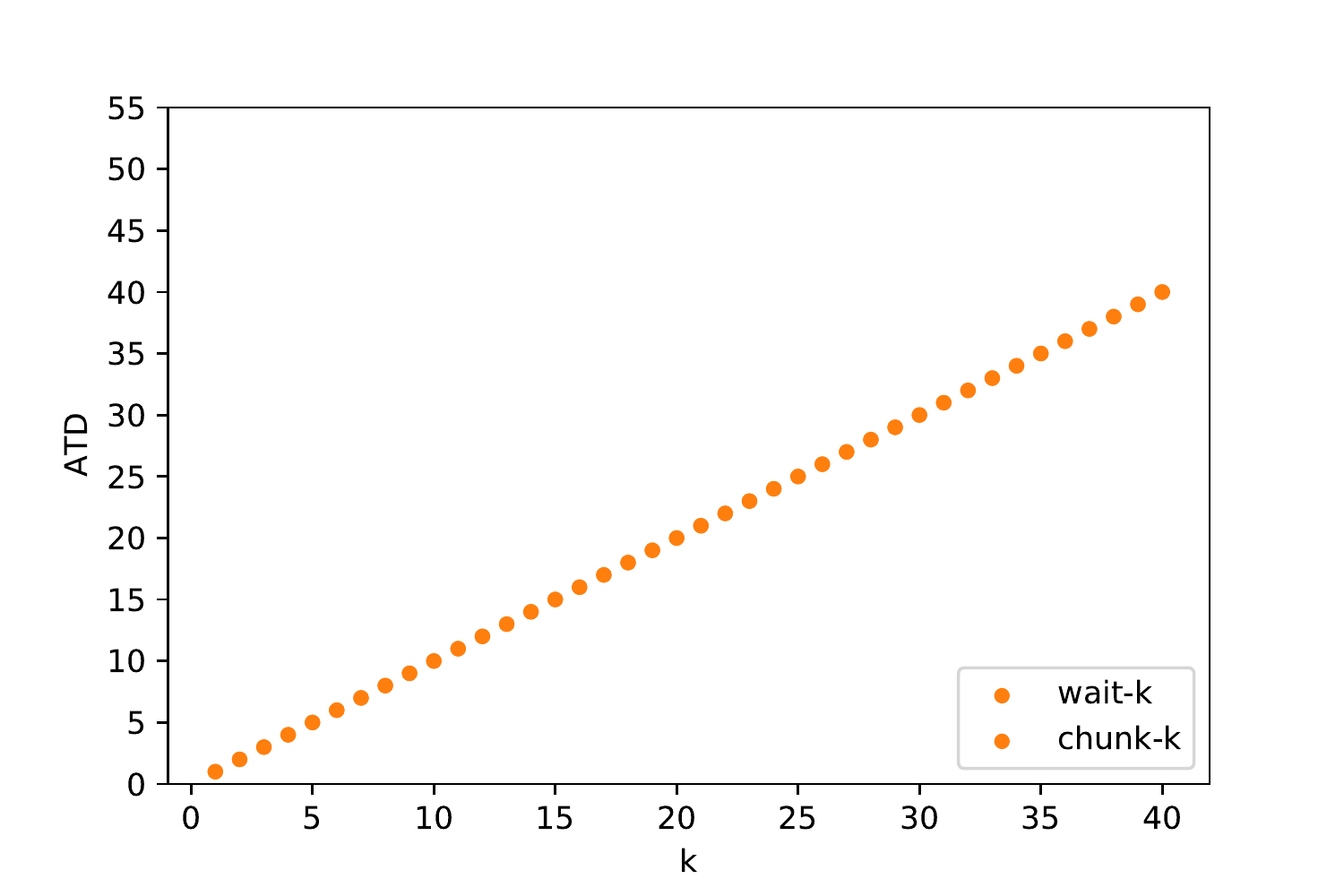}}
\subcaption{Latency measurement by ATD (40-20)}\label{fig:ATDsrc40tgt20}
\end{minipage}
\caption{Case 3 (40 input and 20 output tokens)}\label{fig:case3}
\end{figure}

We also tried a situation with a shorter output of 20 tokens, keeping the input of 40 tokens.
\autoref{fig:case3} shows the results.
An important finding from \autoref{fig:ALsrc40tgt20} is that AL gives negative values both for wait-$k$ and chunk-$k$ with small values of $k$.
AL also suffers from non-smooth changes in latency values even though they are almost continuous unlike \autoref{fig:ALsrc40tgt100}.
ATD results in stable latency measurement both for wait-$k$ and chunk-$k$ in this situation, as shown in \autoref{fig:ATDsrc40tgt20}.

\subsection{Case 4: (20+20)-($L_1$+20)}\label{subsec:case4}

\begin{figure}[t]
\centering
\centerline{\includegraphics[width=8.5cm]{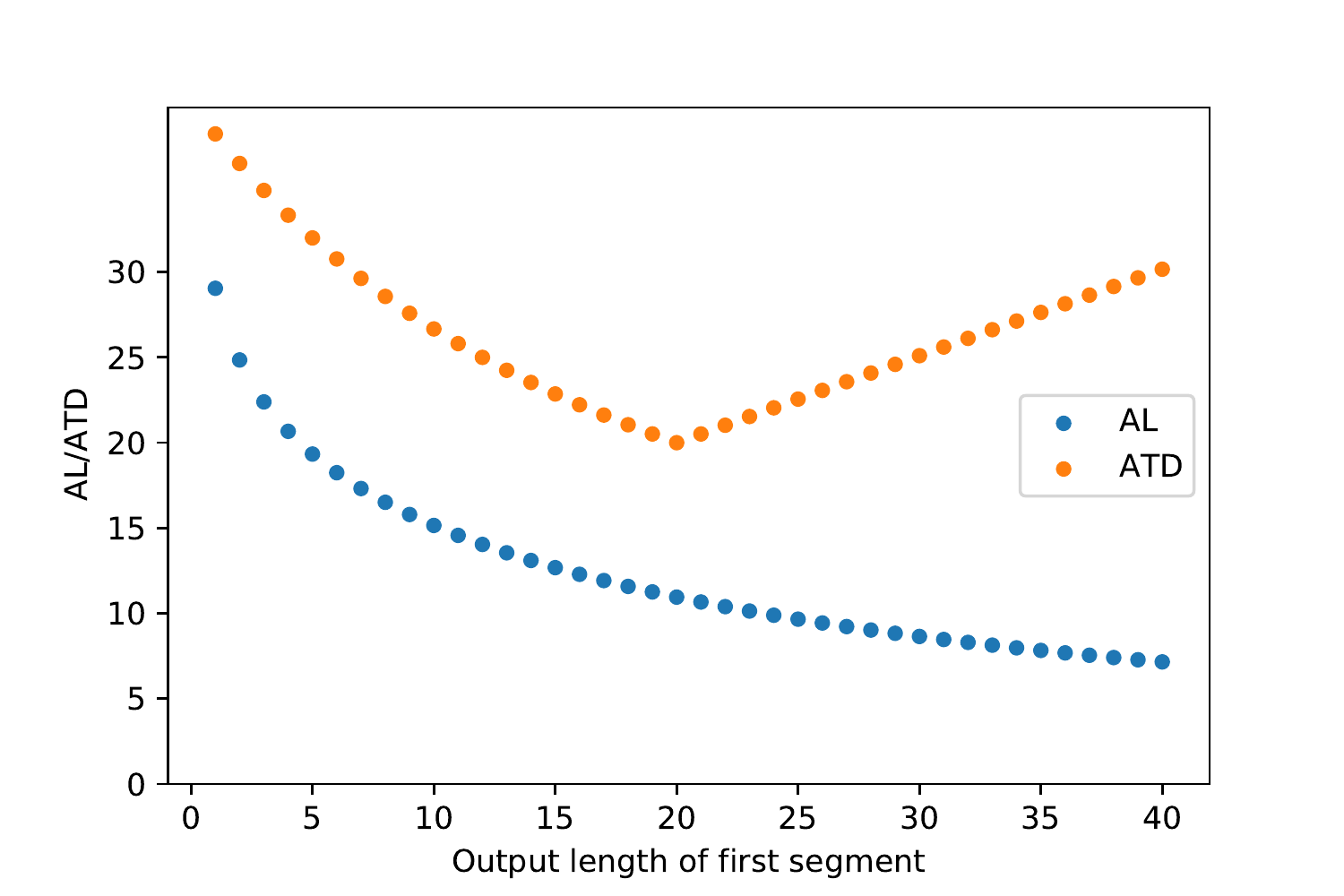}}
\caption{Case 4 (20+20 input and $L_1$+20 output tokens with varying $L_1$)}
\label{fig:ALATD1stSegmentLong}
\end{figure}

We conduct another type of simulation by dividing an input into two parts and varying the length of the output corresponding to the first input part.
Here, we assume the input is divided into two segments with 20 tokens each and the corresponding outputs are with $L_1$ and 20 tokens each with varying $L_1$.
\autoref{fig:ALATD1stSegmentLong} shows the results using a chunk-$20$ strategy.

If $L_1 < 20$, some target tokens corresponding to the first input chunk come after reading the second chunk. 
\autoref{tab:equation1c} shows such situation, in which $L_1 = L_{acc}(y^1) = 1 < 2$. If $y_2$ is in the first chunk output and $L_1 = L_{acc}(y^1) = 2$, the time difference between $x_2$ and $y_2$ becomes  smaller. It also makes the time difference of the latter token pairs such as  $x_3$ and $y_3$ smaller.
Therefore, the shorter $L_1$ becomes, the larger the delay becomes in this range. 

If $L_1 > 20$, the translation of the second chunk is delayed due to the long translation outputs for the first chunk as described in \autoref{tab:equation1b}.
Therefore, the longer $L_1$ becomes, the larger the delay becomes in this range. ATD reflects this phenomena while AL decreases monotonically with $L_1$.


\subsection{Case 5: (20+20)-(20+$L_2$)}\label{subsec:case5}
\begin{figure}[t]
\centering
\centerline{\includegraphics[width=8.5cm]{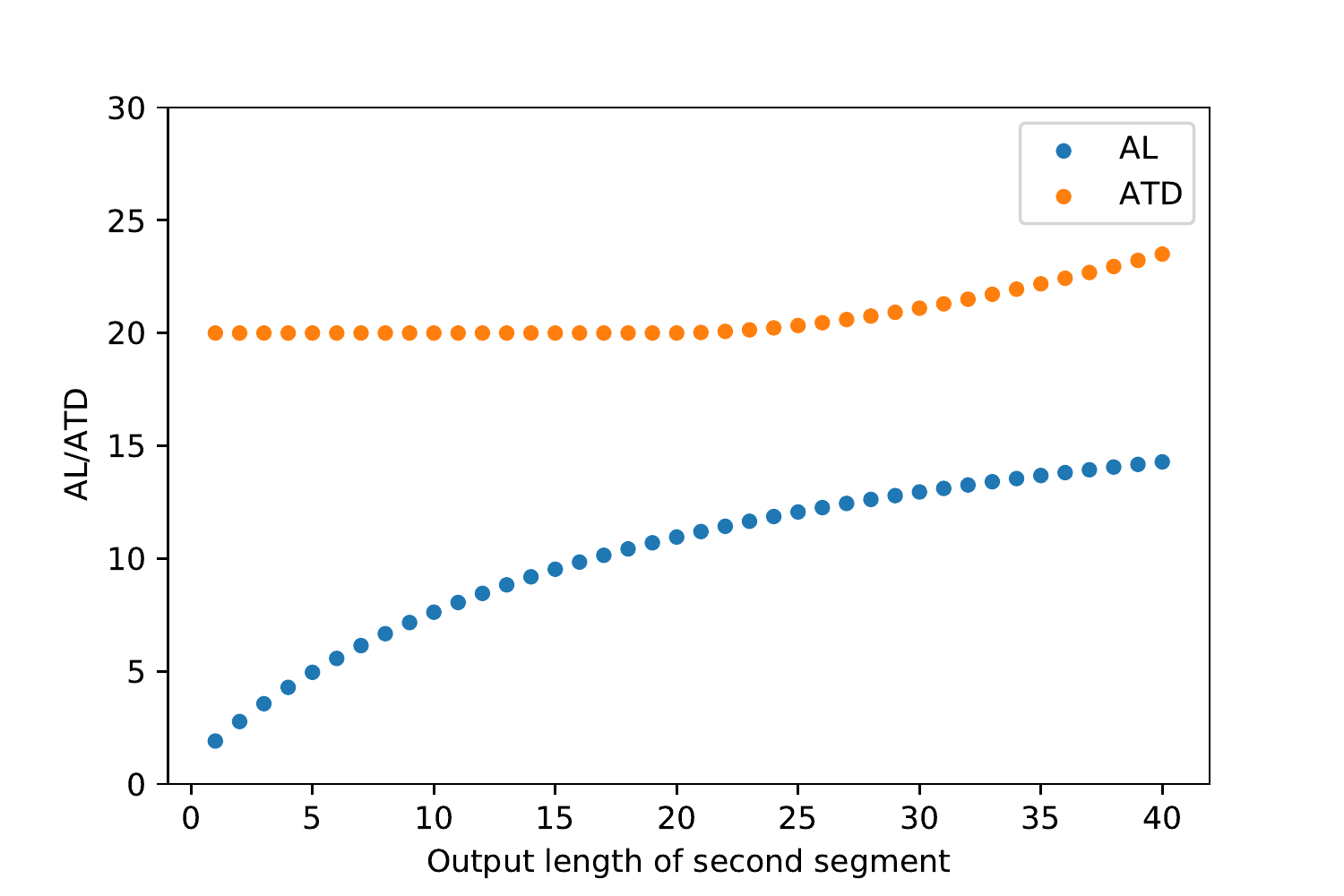}}
\caption{Case 5 (20+20 input and 20+$L_2$ output tokens with varying $L_2$)}
\label{fig:ALATD2ndSegmentLong}
\end{figure}

Then, we analyze a similar situation varying the length of the second output part $L_2$.
Our assumption here is: the input is the same two 20-token segments, and the output consists of the first 20-token segment and the second $L_2$ token segment.
\autoref{fig:ALATD2ndSegmentLong} shows the results using a chunk-$20$ strategy.
Although both AL and ATD values increase with larger $L_2$, AL shows very small values close to zero even though we have to wait for 20 tokens in the first output part;
this is not intuitive as the latency metric.
In contrast, ATD shows a gradual increase in latency values similar to \autoref{fig:ALATD1stSegmentLong} due to additional delays by longer outputs.

\subsection{Case 6: Translation example}\label{subsec:case6}
\begin{figure}[t]
\centering
\centerline{\includegraphics[width=7.5cm]{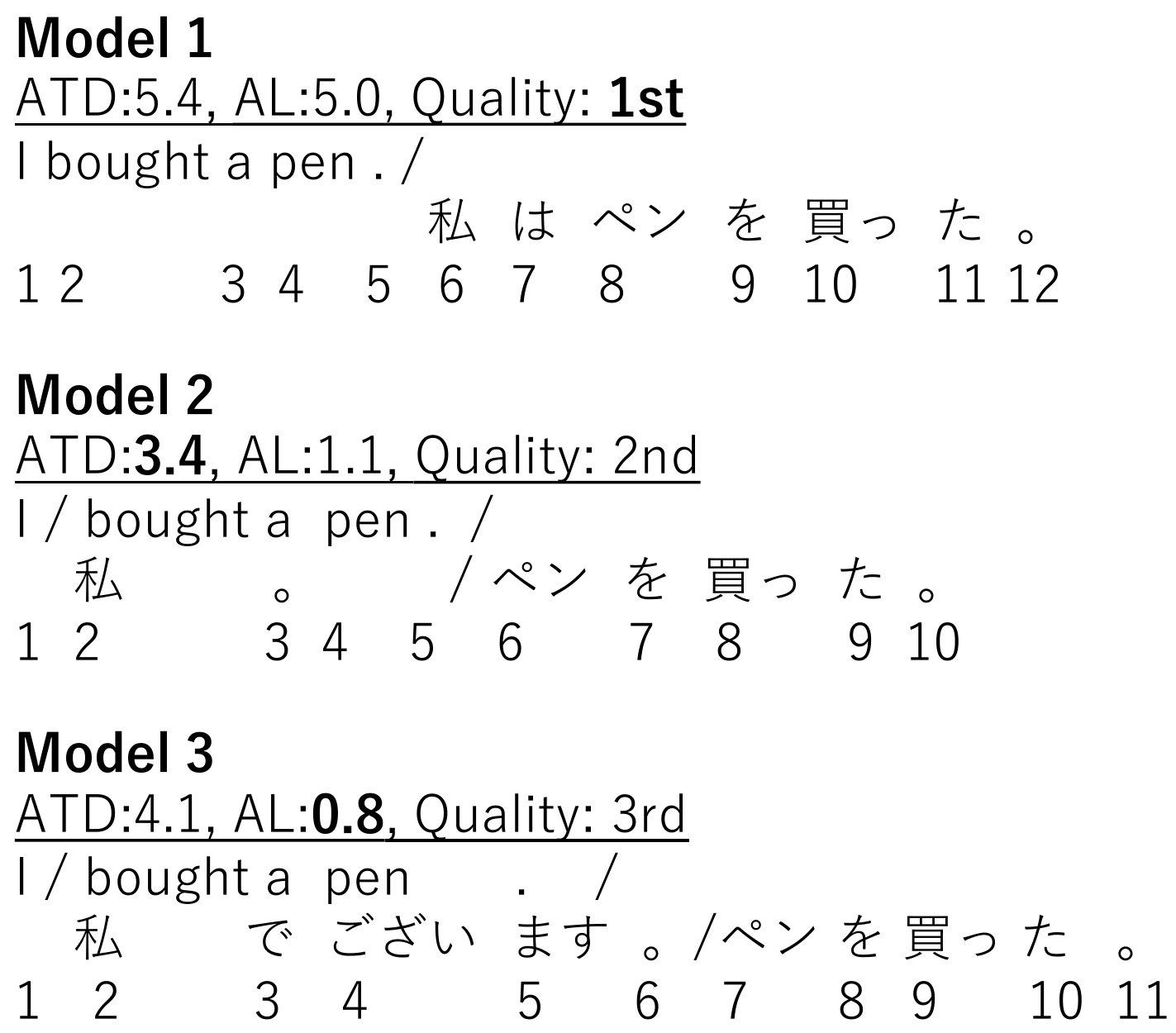}}
\caption{Case 6}
\label{fig:case6}
\end{figure}

\autoref{fig:case6} shows examples of chunk-based simultaneous translation for the input \emph{I bought a pen.} by three different models.

Model 1 waits for the whole input sentence and results in the largest delay and the highest translation quality.

Model 2 has a smaller delay than Model 1 because it can segment the input after observing \emph{I}.
The segmentation enables the model to generate a partial translation but causes quality degradation due to the lack of context.

Model 3 works similarly to Model 2 for the input segmentation but outputs a longer segment for \emph{I}.
This causes large quality degradation by over-translation.

Regarding the latency of these three models, AL is the smallest for Model 3 and decreases largely from Model 1 to Model 2; they are not intuitive as discussed so far.
In contrast, ATD results in reasonable latency values considering the delay caused by the outputs.
\section{Analyses}
We conducted analyses on actual simultaneous translation results to investigate the effectiveness of ATD.
We chose the language directions of English-to-German (en-de) and English-to-Japanese (en-ja) for the comparison of AL and ATD in different conditions in terms of word order.
Note that the analyses here were conducted on simultaneous text-to-text translation for simplicity.

\subsection{Data}
We used the data from the IWSLT evaluation campaign.
For English-to-Geman, we used WMT 2014 training set (4.5 M sentence pairs) for pre-training and IWSLT 2017 training set (206 K sentence pairs) for fine-tuning. 
The development set consists of dev2010, tst2010, tst2011 and tst2012 (5,589 sentence pairs in total), and the evaluation set is tst2015 (1,080 sentence pairs).
For English-to-Japanese, we used WMT 2020 training set (17.9 M sentence pairs) for pre-training and IWSLT 2017 (223 K sentence pairs) for fine-tuning. 
The development set consists of dev2010, tst2011, tst2012, and tst2013 (5,312 sentence pairs in total), and the evaluation set is dev2021 (1,442 sentence pairs).

We compared wait-$k$ \citep{ma-etal-2019-stacl}, Meaningful Unit \citep[MU;][]{zhang-etal-2020-learning-adaptive}, Incremental Constitute Label Prediction \citep[ICLP;][]{kano-etal-2021-simultaneous}, and Prefix Alignment \citep[PA;][]{kano-etal-2022-simultaneous}, following the the experimental settings in the literature \citep{kano-etal-2022-simultaneous}.

\subsection{Results}

\subsubsection{English-to-German}
\begin{figure}[ht]
\begin{minipage}[b]{1.0\hsize}
\centering
\centerline{\includegraphics[width=8.0cm]{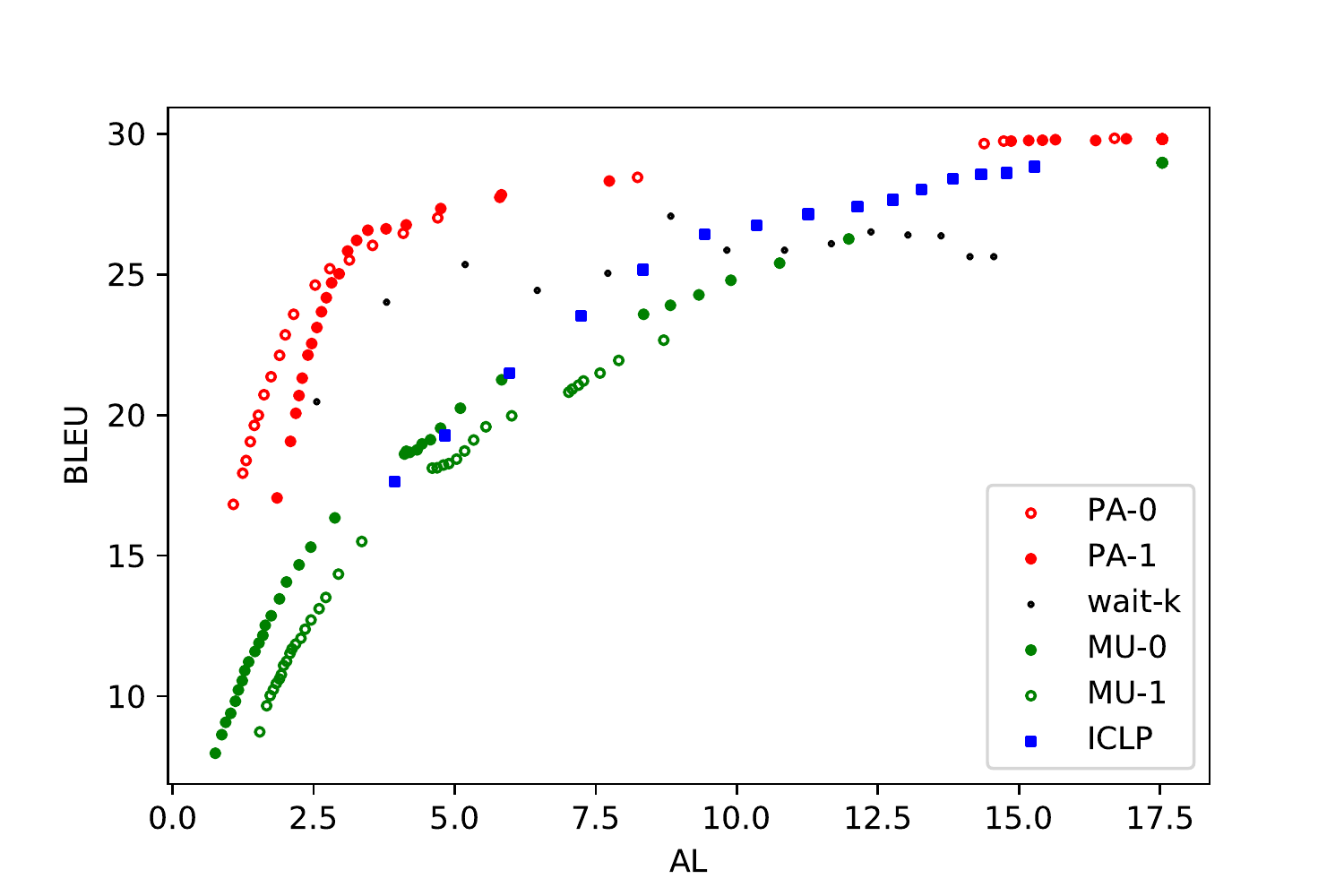}}
\vspace{-2mm}
\subcaption{Latency measurement by AL (en-de)}\label{fig:result_de_al}
\end{minipage}

\begin{minipage}[b]{1.0\hsize}
\centering
\centerline{\includegraphics[width=8.0cm]{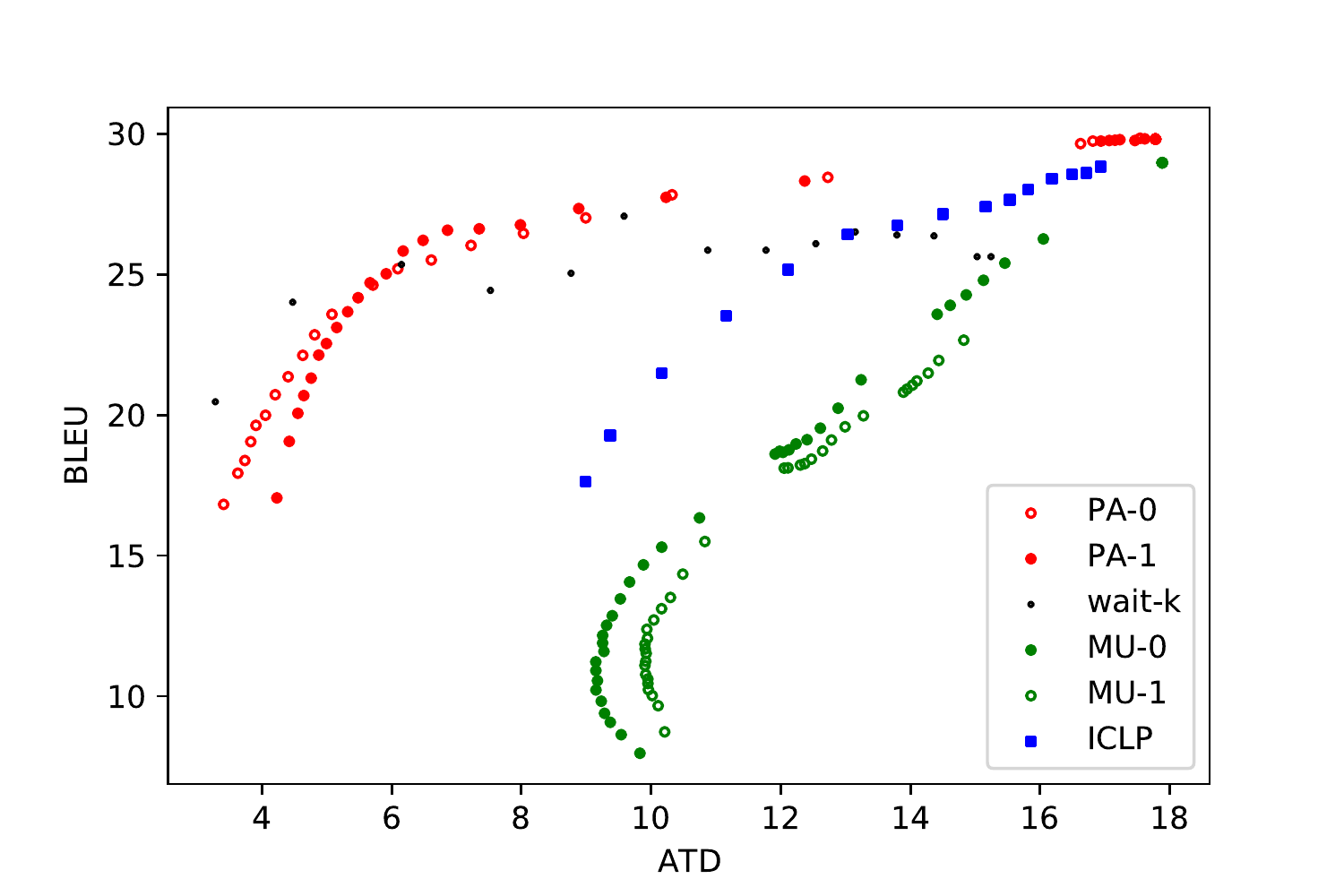}}
\vspace{-2mm}
\subcaption{Latency measurement by ATD (en-de)}\label{fig:result_de_atd}
\end{minipage}
\caption{English-to-German}\label{fig:en-de}

\centerline{\includegraphics[width=8.0cm]{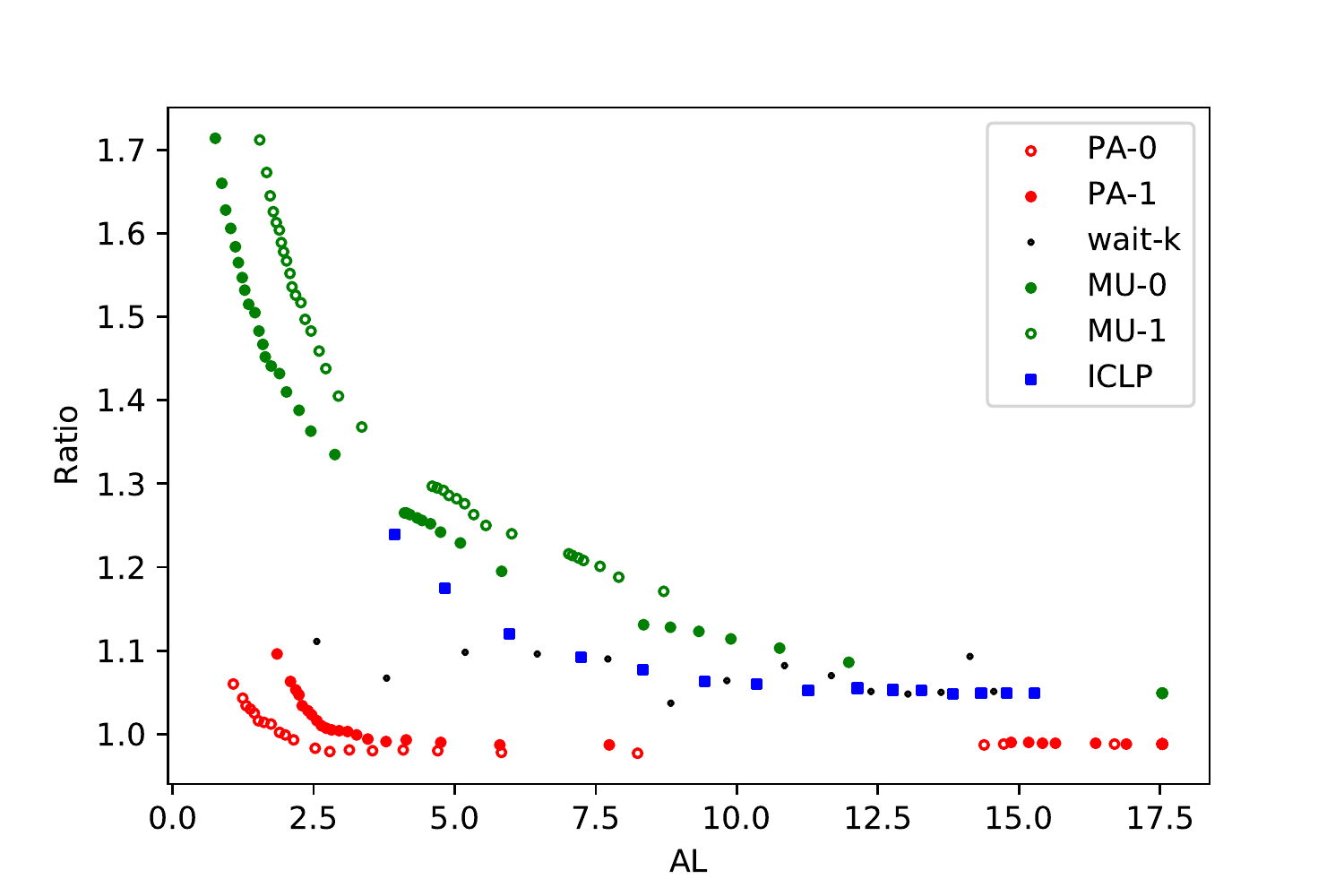}}
\vspace{-4mm}
\caption{Length ratio and AL (en-de)}\label{fig:ratio_de_al}
\end{figure}

\autoref{fig:en-de} shows the results in English-to-German.
Compared to AL shown in \autoref{fig:result_de_al}, ATD shown in \autoref{fig:result_de_atd} demonstrated clear differences in delay among models.
MU and ICLP were affected by the change in the latency metric.
We analyzed their results in detail and found this degradation was due to over-translation as suggested by the observations of length ratio results shown in \autoref{fig:ratio_de_al}.
This phenomenon is the same as what happened with Model 3 in \autoref{subsec:case6}.
MU and ICLP generated long translations exceeding the length ratio of 1.0 when they worked with small latency.
One interesting finding here is the correlation between BLEU and ATD by MU;
larger latency did not always result in better BLEU. It is because over-translation increases ATD, but decreases BLEU at the same time.
In contrast, wait-$k$ is a strategy that generates one output token at a time and can avoid such an issue.
PA also worked well with the latency measurement by ATD because it fine-tunes the translation model to avoid over-translation.

\subsubsection{English-to-Japanese}

\begin{figure}[t]
\begin{minipage}[b]{1.0\hsize}
\centering
\centerline{\includegraphics[width=8.0cm]{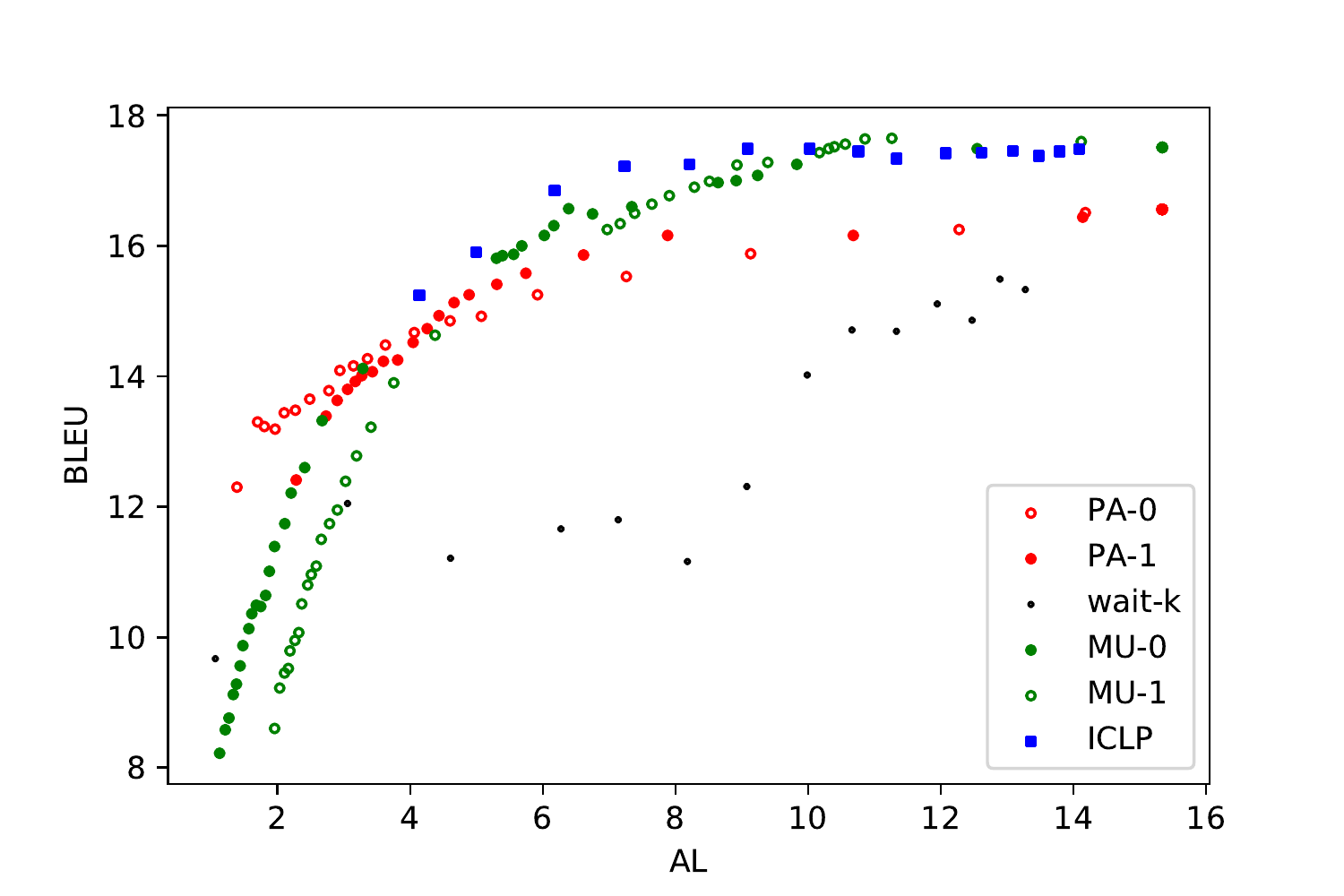}}
\vspace{-2mm}
\subcaption{Latency measurement by AL (en-ja)}\label{fig:result_ja_al}
\end{minipage}
\begin{minipage}[b]{1.0\hsize}
\centering
\centerline{\includegraphics[width=8.0cm]{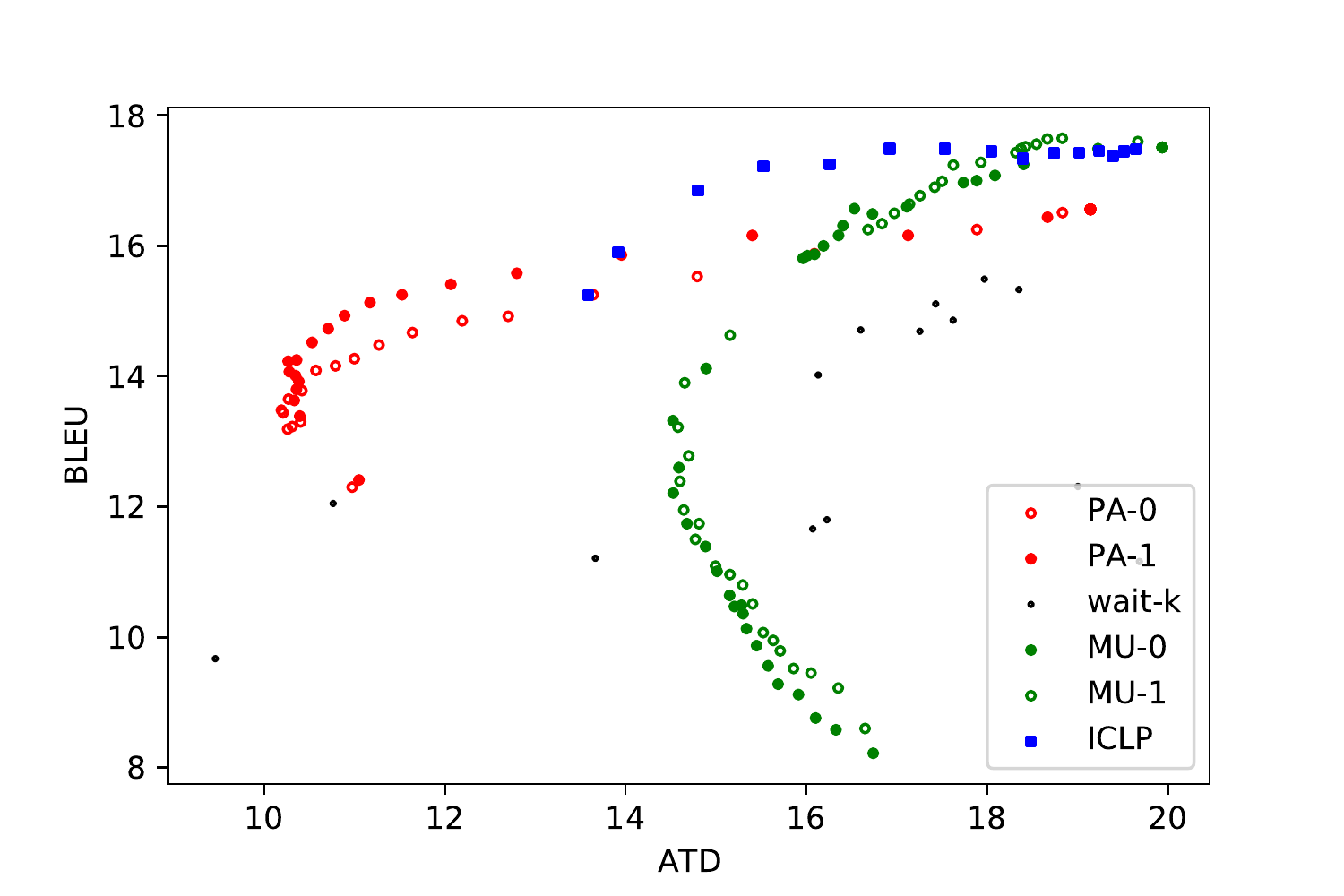}}
\vspace{-2mm}
\subcaption{Latency measurement by ATD (en-ja)}\label{fig:result_ja_atd}
\end{minipage}
\caption{English-to-Japanese}\label{fig:en-ja}
\end{figure}

\autoref{fig:en-ja} shows the results in English-to-Japanese.
The results were similar, but the actual latency values of ATD were much larger than those in English-to-German.
This is also due to the limitation of the current implementation of ATD based on SimulEval \cite{ma-etal-2020-simuleval}, where the latency in text-to-text English-to-Japanese translation is calculated based on the number of Japanese characters, not the number of words as in English-to-German.
Therefore, the number of output tokens had to be much larger than the number of input tokens in English-Japanese, so the ATD value became large.
This large number of output tokens made the effect of over-translation by MU in \autoref{fig:result_ja_atd} more serious than that in English-to-German.

\section{Conclusion}

We proposed a novel latency metric ATD for simultaneous machine translation, which addresses the problem in the latency evaluation for a chunk-based model by taking the output length into account.
ATD gives a large latency value to a long output based on the assumption that the output also causes a delay, different from AL.
We revealed the effectiveness of ATD through the analyses of simulation and actual translation results compared with AL.

Future work includes studies on semantics-oriented latency measurement not just focusing on timing information without any consideration about the delivery of contents.



\section*{Acknowledgements}
 Part of this work is supported by JSPS KAKENHI Grant Number JP21H05054.

\bibliography{anthology,custom}




\end{document}